\documentclass[twoside,11pt]{article}

\usepackage{jmlr2e}
\usepackage{times}
\usepackage{latexsym}
\usepackage{url}
\usepackage{helvet}  
\usepackage{courier}
\usepackage{graphicx}
\usepackage{booktabs} % Allows the use of \toprule, \midrule and \bottomrule in tables
\usepackage{color}
\usepackage{multirow}
\usepackage{bbm}
\usepackage{hyperref}
\usepackage[ruled]{algorithm2e}
\usepackage{float}
\usepackage{caption}
\usepackage{subfig}
\usepackage{wrapfig}
\usepackage[T1]{fontenc}
\usepackage[utf8]{inputenc}
\usepackage{amsmath}
\usepackage{amsfonts}
\usepackage{lipsum}

\newcommand{\Gmat}{{\bf G}}

\newcommand{\Wmat}[0]{{{\bf W}}}
\newcommand{\Xmat}[0]{{{\bf X}}}

\newcommand{\bv}[0]{{\boldsymbol{b}}}
\newcommand{\cv}[0]{{\boldsymbol{c}}}

\newcommand{\ev}[0]{{\boldsymbol{e}}\xspace}

\newcommand{\gv}[0]{{\boldsymbol{g}}\xspace}
\newcommand{\hv}[0]{{\boldsymbol{h}}}

\newcommand{\qv}[0]{{\boldsymbol{q}}\xspace}

\newcommand{\xv}{\boldsymbol{x}}
\newcommand{\yv}{\boldsymbol{y}}
\newcommand{\zv}{\boldsymbol{z}}

\newcommand{\Thetamat}{\boldsymbol{\Theta}}

\newcommand{\Lambdamat}{\boldsymbol{\Lambda}}

\newcommand{\epsilonv}{\boldsymbol{\epsilon}}

\ShortHeadings{Multi-Label Learning from Medical Plain Text with Convolutional Residual Models}{Zhang, Henao, Gan, Li, \& Carin}
\firstpageno{1}

\begin{document}

\title{Multi-Label Learning from Medical Plain Text with Convolutional Residual Models}

\author{\name Xinyuan Zhang \email xy.zhang@duke.edu \\
       \addr Department of Electrical and Computer Engineering\\
       Duke University\\
       Durham, NC, USA
       \AND
       \name Ricardo Henao \email ricardo.henao@duke.edu \\
       \addr Department of Biostatistics and Bioinformatics\\
       Duke University\\
       Durham, NC, USA
   	   \AND
       \name Zhe Gan \email zhe.gan@duke.edu \\
       \addr Department of Electrical and Computer Engineering\\
       Duke University\\
       Durham, NC, USA
       \AND
       \name Yitong Li \email yitong.li@duke.edu \\
       \addr Department of Electrical and Computer Engineering\\
       Duke University\\
       Durham, NC, USA
       \AND
       \name Lawrence Carin \email lcarin@duke.edu \\
       \addr Department of Electrical and Computer Engineering\\
       Duke University\\
       Durham, NC, USA} 

\maketitle

\begin{abstract}
Predicting diagnoses from Electronic Health Records (EHRs) is an important medical application of multi-label learning.
We propose a convolutional residual model for multi-label classification from \emph{doctor notes} in EHR data. A given patient may have multiple diagnoses, and therefore multi-label learning is required.
We employ a Convolutional Neural Network (CNN) to encode plain text into a fixed-length sentence embedding vector.
Since diagnoses are typically correlated, a deep residual network is employed on top of the CNN encoder, to capture label (diagnosis) dependencies and incorporate information directly from the encoded sentence vector.
A real EHR dataset is considered, and we compare the proposed model with several well-known baselines, to predict diagnoses based on doctor notes. Experimental results demonstrate the superiority of the proposed convolutional residual model.  
\end{abstract}

\section{Introduction}
Machine learning is playing an increasingly important role in medical applications.
The widespread use of Electronic Health Records (EHRs) has made it possible to collect massive amounts of data from patients and health providers.
Machine learning technology can help clinical researchers ($i$) discover hidden patterns from massive EHR data, and ($ii$) develop predictive models to assist with clinical decision making.
One important application of multi-label learning in the medical domain is to predict diagnoses given features from the EHR.

Multi-label learning is a supervised classification framework in which multiple target labels can be assigned to one instance.
For example, in our motivating doctor notes dataset, one patient can be associated with multiple diagnoses simultaneously (comorbidities), {\emph e.g.}, ``fever'', ``cough'', and ``viral infection'', as one patient may suffer from several related illnesses.
One challenge of multi-label learning is modeling label dependencies by realizing that labels are often correlated.
In the above example, and assuming that the $\{$fever,~cough,~viral infection$\}$ combination is likely, if a patient manifests fever and cough, then the probability of also having a viral infection is likely to increase accordingly.

Traditional multi-label learning methods such as \emph{one-versus-all} and \emph{one-versus-one} \cite{zhang2014review} assume that labels are independent of each other.
Recent work in multi-label learning focuses on exploiting label correlation to improve classification performance.
A natural approach consists of using embedding-based models to project label vectors onto a low dimensional space while capturing dependencies, thus reducing the ``effective'' number of labels \cite{bhatia2015sparse}. However, in practice, embedding-based methods can result in subpar performance due to the loss of information during the embedding procedure.
Tree-based approaches achieve faster prediction by recursively partitioning labels into tree-structured groups \cite{prabhu2014fastxml}. However, errors made at the upper levels of the hierarchy cannot be corrected at bottom levels, which often leads to a loss in overall predictive accuracy.
In addition, most models referenced above use \emph{bag-of-words} sentence representations.
The main shortcoming of bag-of-words models is their inability to capture local context (semantics).
For example, ``buy used cars'' and ``purchase old automobiles'' are represented by orthogonal (unrelated) vectors in a bag-of-words representation, but in fact they are semantically identical.
The order of words is not respected in bag-of-words representations.
For example, ``he is healthy'' and ``is he healthy'' have exactly the same bag-of-words representation but have vastly different meanings.
Long Short-Term Memory (LSTM) \cite{hochreiter1997long} is a widely used approach to estimate fixed-length vectorial representations to capture sentence meaning. However, an LSTM typically becomes ineffective when modeling very long sentences or paragraphs, which unfortunately are typical in doctor notes.
%due to exposure bias \cite{bengio2015scheduled}.
%Unfortunately, long sentences or paragraphs are typical in doctor notes.

In this paper we consider a real EHR-based dataset, where the goal is to predict diagnoses based on plain-text doctor notes.
This scenario is not a typical sentence classification task, in the sense that each note is composed of several sentences (a paragraph).
The average note length is $101\pm48$ words.
To evaluate the proposed model, we compare it with a number of related methods on our doctor notes dataset.
Experiments show that our convolutional residual model outperforms all other competing methods.
The superiority of the proposed model indicates that: ($i$) compared to bag-of-words models, the CNN is more effective at sentence encoding, by leveraging semantics and word ordering, and ($ii$) deep residual networks successfully capture label dependencies, thus delivering significantly improved multi-label classification performance.

\paragraph{Technical Significance}
We develop a convolutional residual model to address this multi-label plain text learning problem.
CNN models have been increasingly used for Natural Language Processing (NLP) applications, and have achieved excellent results on both supervised \cite{kalchbrenner2014convolutional,hu2014convolutional} and unsupervised \cite{gan2017learning} tasks.
Here, we employ a CNN as the sentence (doctor's note) encoder, due to its excellent performance at identifying sentence structure, especially in long noisy sentences.
A deep residual network \cite{he2016deep} is added on top of the CNN encoder, to capture label correlations and incorporate information from the encoded sentence vector (CNN's output).
This is achieved using shortcut connections between layers \cite{bishop1995neural,venables2013modern}.
A common problem in deep networks is that performance saturates, then degrades rapidly as the network grows deeper.
Deep residual networks mitigate this problem by having stacked layers that fit a residual mapping.

\paragraph{Clinical Relevance}
This work focuses on predicting diagnoses given a plain doctor's note on a patient's presentation, symptoms, medical history, etc.
However, leveraging the dependencies of multiple comorbidities in an heterogeneous population of patients remains a very challenging problem.
Exploiting diagnoses correlation (co-occurrence) is essential when approaching this predictive problem with multi-label classification methods.
Further, a large number of typos, medical jargon, and non-standard abbreviations make the notes considerably noisy and heterogeneous.
Our goal is to build a classifier as robust as possible while minimizing the preprocessing burden, to address the above problems and simplify the implementation of the predictive model in our local Medical System.

\section{Related Work}
There is an extensive literature on applying NLP techniques to medical-domain tasks.
Notably, \cite{jagannatha2016structured} applied RNN-based sequence labeling in phrase detection of medical text.
\cite{kuo2016ensembles} built an NLP ensemble pipeline to integrate two systems cTAKES \cite{savova2010mayo} and MetaMap \cite{aronson2010overview} for biomedical data-element extraction from clinical text.
Recently, \cite{li2017convolutional} employed CNN on textual admission information to predict medical diagnosis.
However, none of these approaches consider the (practical) problem of diagnosing patients affected by multiple illnesses, which belongs to the larger class of multi-label learning methods.

Existing multi-label classification methods are commonly grouped into two broad categories.
One is problem-transformation methods, that convert a multi-label learning problem into several binary classification tasks by using existing techniques such as Binary Relevance (BR) \cite{tsoumakas2009mining}, Label Power-set (LP) \cite{tsoumakas2006multi} and the pair-wise method \cite{wu2004probability}.
The other category consists of algorithm-adaptation methods, which extend specific supervised algorithms to deal with multi-label data.
For instance, \cite{li2015conditional} leveraged the conditional Restricted Boltzmann Machine (RBM) to multi-label tasks.
Our model belongs to the latter group, in the sense that we extend CNN-based classifiers.

Recently, multi-label learning has been employed for predicting diagnoses based on clinical data.
\cite{zhao2013clinical} developed the Ensemble of Sampled Classifier Chains (ESCC) algorithm by exploiting disease label relationships, to classify clinical text according to their specified disease labels.
\cite{li2016multi} performed multi-label classification for health and disease risk prediction using a Multi-Label Problem Transformation Joint Classification (MLPTJC) method.
However, none of these methods use plain text data nor CNN-based representations.

\section{Methods}
We use CNNs as a fixed-length feature extractor from word sequences, {\emph i.e.}, doctor notes in our motivating dataset.
Doctor notes are in general composed of several sentences. Here we treat each note as a single ``meta'' sentence, with end-of-sentence tokens located at the end of each actual sentence. 
Three different classifiers based on the CNN-encoded sentence features are described and evaluated, including our convolutional residual model described in Section~\ref{sec: conv_residual}.

A sentence $S$ with $T$ words $\{w_1,w_2,...,w_T\}$ is mapped into a set of $k$-dimensional real-valued vectors $\{\ev_1,\ev_2,...,\ev_T\}$ using a word embedding.
Word vectors are initialized using \textit{word2vec} which is pre-trained on $100$ billion words from Google News, including numbers and special characters, using a continuous bag-of-words architecture \cite{mikolov2013distributed}. This embedding matrix is further refined using medical text during training.

\subsection{CNN Sentence Encoder}\label{sc:cnn_enc}
\begin{wrapfigure}{R}{0.5\textwidth}	
	\centering
	\includegraphics[width=0.36\textwidth]{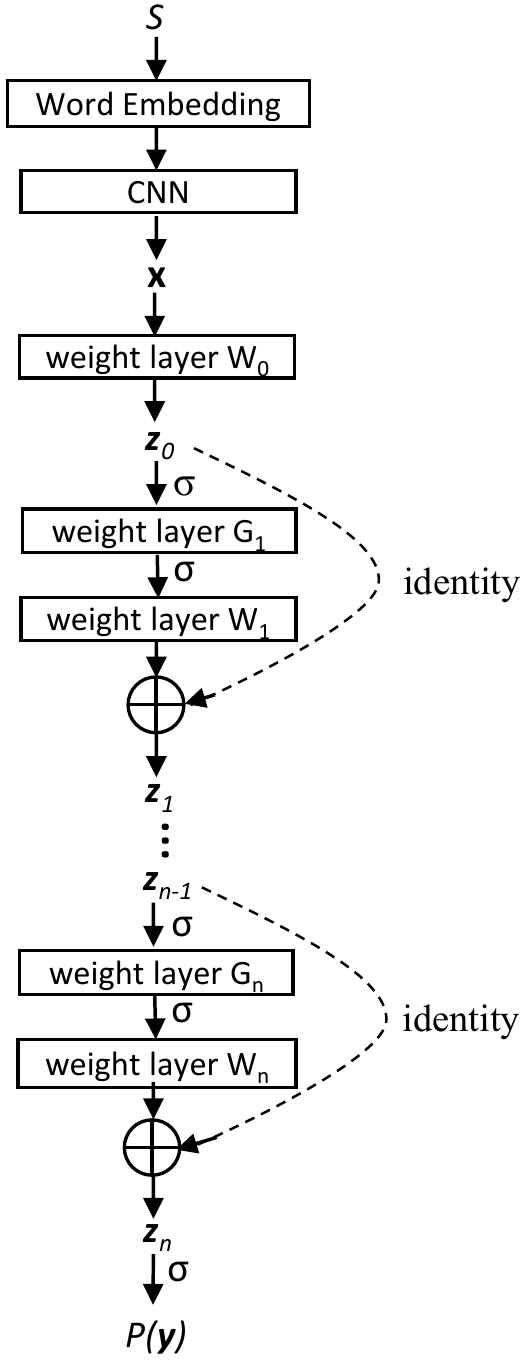}
	\caption{The building blocks of an $n$-layer convolutional residual model.}
	\label{fig:blocks2}
	\vspace{5mm}
\end{wrapfigure}

Based on the word embedding, sentence $S$ is represented as a $\Xmat\in\mathbb{R}^{k\times T}$ matrix, assembled by concatenating word vectors $\ev_1$, $\ev_2$,..., $\ev_T$, {\emph i.e.},
\begin{align*}
	\Xmat=\ev_1\oplus \ev_2\oplus...\oplus \ev_T \,,
\end{align*}
where the $i$-th column of $\Xmat$ is the embedding vector corresponding to word $w_i$.

The CNN from \cite{kim2014convolutional,collobert2011natural} with $\Xmat$ as input is utilized as the sentence encoder.
Given a filter $\Lambdamat\in\mathbb{R}^{k\times t}$ with a window of $t$ words, we produce feature map $\gv=[g_1\ g_2\ ...\ g_{T-t+1}]$ by
\begin{align}\label{eq:fconv}
	\gv=f(\Xmat*\Lambdamat+\epsilonv) \,,
\end{align}
where $\gv\in\mathbb{R}^{T-t+1}$, $f(\cdot)$ is a nonlinear function such as the hyperbolic tangent used in this paper, $*$ is the convolution operator and $\epsilonv\in\mathbb{R}^{T-t+1}$ is a bias term.

Note the obtained feature map, $\gv$, of length $T-t+1$ depends of the sentence length.
To deal with this issue, the {\emph max-pooling} operator is applied on the feature map.
By taking the maximum value $\hat{g}=\max\{\gv\}$, only the most salient sentence feature corresponding to the filter, $\Lambdamat$, is captured for sentence $S$ via $\Xmat$.

Equation~\eqref{eq:fconv} describes how the model uses one filter to extract one sentence feature.
In practice, multiple filters of varying window sizes act as linguistic feature detectors, whose goal is to recognize specific classes of $n$-\emph{grams}.
For instance, if we set $v$ filters, where each filter has $w$ variant window sizes, then the resulting encoded sentence representations is a $vw$-dimensional vector.

\subsection{CNN Classifiers}\label{cnnclassifier}
%A typical CNN-based binary classifier specifies a logistic regression model on top of the CNN architecture.
Let $\xv\in\mathbb{R}^{vw}$ be the encoded sentence vector.
In multi-label learning, the label layer $\yv=[y^{(1)} \ \ldots \ y^{(L)}]\in\{0,1\}^L$ represents the underlying true label vector (diagnoses in our case), where $L$ is the total number of labels and $y_i=1$ indicates the existence of $i$-th label.
Given a sentence vector $\xv$, the probability of all labels $P(\yv)=[P(y^{(1)}=1) \ \ldots \ P(y^{(L)}=1)]$ can be expressed as
\begin{align}\label{eqt:class}
	P(\yv)=\sigma(\zv_0) \,, \quad \zv_0 = \Wmat_0\xv + \bv_0 \,,
\end{align}
where $P(\yv)\in[0,1]^L$, $\sigma(\cdot)$ denotes the sigmoid link function, $\sigma(x)=1/(1+\exp(-x))$, $\Wmat_0\in\mathbb{R}^{L\times vw}$ is the (classification) weights matrix, and $\bv_0\in\mathbb{R}^L$ is the bias term.

\subsection{Conditional RBM Models}\label{sec-CRBM}
The CNN classifier in \eqref{eqt:class} treats each label independently, as $L$ binary classification problems.
However, in real multi-label tasks the labels are usually correlated with each other.
The model proposed by \cite{li2015conditional} for bag-of-words representations uses a Restricted Boltzmann Machine (RBM) to capture this high-order label dependency.

A latent layer $\hv\in\{0,1\}^J$ with $J$ hidden units is added above the label layer $\yv\in\{0,1\}^L$.
Conditioning on the input feature layer $\xv$, the encoded sentence vector in our case, layers $\yv$ and $\hv$ form a standard restricted Boltzmann machine.
The Conditional RBM (CRBM) model is specified via parameters: $\Wmat\in\mathbb{R}^{L\times vw},\Gmat\in\mathbb{R}^{L\times J},\bv\in\mathbb{R}^{L\times 1},\cv\in\mathbb{R}^{J\times 1}$.
The conditional marginal likelihood is defined as
\begin{align*}%\label{CRBM}
	P(\yv|\xv)=\frac{\sum_{h}\exp[-E_{\rm con}(\yv,\xv)-E_{\rm rbm}(\yv,\hv)]}{Z(\xv)} \,,
\end{align*}
where $Z(\xv)$ is the normalization factor, and $E_{\rm con}(\yv,\xv)$ and $E_{\rm rbm}(\yv,\xv)$ are two energy functions such that
\begin{align*}
	& Z(\xv)=\sum_{\yv}\sum_{\hv}\exp[-E_{\rm con}(\yv,\xv)-E_{\rm rbm}(\yv,\hv)] \,, \\
	& E_{\rm con}(\yv,\xv)=-\yv^\top \Wmat\xv \,, \\
	& E_{\rm rbm}(\yv,\hv)=-\yv^\top \Gmat\hv-\yv^\top \bv-\cv^\top \hv \,.
\end{align*}
Given training data, the CRBM model is optimized by maximizing the conditional marginal likelihood, $P(\yv|\xv)$.

According to the conditional independence structure of RBM models, the local conditional probabilities can be computed as
\begin{align}\label{eq:rbm_cond}
	\begin{aligned}
		P(h^{(j)} = 1|\yv,\xv) = & \sigma(\yv^\top \Gmat_{:j}+\cv_j) \,, \\
		P(y^{(l)} = 1|\hv,\xv) = & \sigma(\Gmat_{l:}\hv+\bv_{l}+\Wmat_{l:}\xv) \,.
	\end{aligned}
\end{align}
where $\Gmat_{:j}$ and $\Gmat_{l:}$ are column $j$ and row $l$ of $\Gmat$, respectively.
%
%
%\begin{figure}[t!]
%	\centering 
%	\includegraphics[width=1.5in]{fig2.pdf} 
%	\caption{The building blocks of an $n$-layer convolutional residual model.}
%	\label{fig:blocks2}
%\end{figure}

\subsection{Convolutional Residual Models} \label{sec: conv_residual}

Motivated by the local conditional probabilities of CRBM models in \eqref{eq:rbm_cond}, we add feedforward neural networks with shortcut connections on top of the CNN encoder in Section \ref{sc:cnn_enc}.
Shortcut connections are defined as those skipping one or more layers.
The idea is to incorporate information from both the sentence layer via the encoded sentence vector, $\xv$, and the label layer, $\yv$, the latter to capture label dependencies.
These two components form essentially a \emph{deep residual network}.
Shortcut connections capture the predictive interactions between the encoded sentence vectors and the output labels, while the stacked feedforward neural network captures the correlations in the label layer.
Shortcut connections enable the stacked layers to fit a residual mapping, which avoids model degradation as the network depth (number of layers) increases \cite{he2016deep}.
In this paper, shortcut connections are constructed as identity mappings, thus the encoded sentence vectors are directly added to the outputs of the stacked layers.
\begin{algorithm}[t!]
	\SetKwInOut{Input}{Input}
	\SetKwInOut{Output}{Output}
	\Input{Sentence $S$ with $T$ words.}
	\Output{$P(\yv)$: Probability of each label.}
	Encode $S$ into vector $\xv\in\mathbb{R}^{vw}$ using CNN.\\
	Initialize base layer: $\zv_0=\Wmat_0\xv+\bv_0$.
	
	% \begin{equation*}
	% \zv_0=\Wmat_0\xv+\bv_0 \,.
	% \end{equation*}
	\For{$i=1,2,...,n$}{
		$\qv_i=\sigma(\zv_{i-1})\Gmat_i+\cv_i$.
		
		$\zv_i=\Wmat_0\xv+\bv_i+\sum_{t=1}^{i}\Wmat_t\sigma(\qv_t)$.
		%\begin{align*}
		%& \qv_i=\sigma(\zv_{i-1})\Gmat_i+\cv_i \,. \\
		% & \zv_i=P(\yv_{i-1})\Gmat_i+\cv_i \,. \\
		% & P(\yv_i)=\sigma\left(\Wmat_0\xv+\bv_i+\sum_{t=1}^{i}\Wmat_t\sigma(\zv_t)\right) \,. \\
		%& \zv_i=\Wmat_0\xv+\bv_i+\sum_{t=1}^{i}\Wmat_t\sigma(\qv_t) \,.
		%\end{align*}
	}
	$P(\yv)$ = $\sigma(\zv_n)$.
	\caption{Convolutional Residual Model.}
	\label{alg: CNN-Res}
\end{algorithm}

%\vspace{-0.5in}
Our complete convolutional residual model is composed of the CNN encoder in Section \ref{sc:cnn_enc} and the residual classifier described above.
The building blocks of an $n$-layer residual classifier are shown in Figure~\ref{fig:blocks2}, in which $\oplus$ denotes element-wise addition.
$\xv$ represents the CNN encoded sentence vector and $\Wmat_0$ is the weight in Equation \ref{eqt:class}.
We set sigmoid link functions, $\sigma(\cdot)$, as nonlinearities on every layer, and the biases are omitted for simplicity.
Further, $n$ is the number of residual layers and $h_1,...,h_i,...,h_n$ are the number of hidden units of each hidden layer.
$\Wmat_0\in\mathbb{R}^{L\times vw}$, $\Wmat_1,...,\Wmat_n\in\mathbb{R}^{L\times h_i}$, $\bv_0,\bv_1,...,\bv_n\in\mathbb{R}^L$, $\Gmat_1,...,\Gmat_n\in\mathbb{R}^{L\times h_i}$, and $\cv_1,...,\cv_n\in\mathbb{R}^{h_i}$ are parameters to be learned, namely, weight matrices and bias vectors for layers $0,\ldots,n$.
Identity shortcut connections introduce neither additional parameters nor computational complexity. This allows us to fairly compare plain stacked classifiers and residual classifiers.
The algorithm for our \textit{Convolutional Residual Model} for multi-label tasks is presented as Algorithm~\ref{alg: CNN-Res}.
A plain stacked classifier (without shortcut connections) correspond to the architecture in Figure~\ref{fig:blocks2} without the identity connections (dashed lines).

The residual and plain stacked classifiers have exactly the same number of parameters.
Let $\Thetamat$ be the set of parameters of both the CNN encoder and the residual classifier.
We wish to find the optimal set of parameters that minimize the cross-entropy loss function, expressed as
\begin{align}
	\arg\min_{\Thetamat} \ \ \frac{1}{L}\sum_{i=1}^{L}y^{(i)}\log P(y^{(i)}=1) \label{cross-entropy}+(1-y^{(i)})\log[1-P(y^{(i)}=1)] \nonumber \,.
\end{align}
The parameters, $\Thetamat$, of the entire network are jointly optimized by stochastic gradient descent with back-propagation.
%\cite{network1989handwritten}.

\section{Doctor Notes Data}\label{sc:data}
This work is motivated by a real EHR cohort collected from patients at Duke Hospital. 
The input plain text is a doctor's note on a patient's history of present illness, which briefly describes in the doctor's words information about the patient with regard to presentation, symptoms, medical history, \emph{etc}.
The outputs (labels) are free-text discharge diagnoses, medications, and dispositions.
In this paper we focus on diagnoses as the target of interest.
Note that each patient could have multiple diagnoses simultaneously.
Furthermore, we do not preprocess the plain text data in any way, which means we retain typos, medical jargon and non-standard abbreviations in the notes.

It is important to consider that discharge diagnoses adjudication does not occur immediately after the doctor's note has been written.
In fact, in our data, about 50\% of the patients have to wait for at least 7 hours to be diagnosed, 25\% have to wait for at least 18 hours and only 12\% of patients have to wait less than 2 hours.
This time is usually spent on laboratory tests and medical procedures required to confirm the discharge diagnoses. Hence, our predictions are of {\em future} diagnoses based on doctor notes, which presents doctors with the most likely diagnoses to guide orders and hopefully improve care.
\begin{table*}[t!]
	\centering  
	\begin{tabular}{l|l|l|l|l|l|l|l}
		Dataset & $L$ & Label Type & Controls & $N$ & $|V|$ & $l$ & $N_V$ \\ 
		\hline
		\textit{ehr25} & 25 & Single-Label & No & 7897 & 19596 & 410 & 717 \\
		\textit{ehr1000} & 1000 & Multi-Label & No & 44473 & 54215 & 600 & 4447 \\
		\textit{ehr64-all} & 64 & Multi-Label & Yes & 50128 & 58359 & 600 & 5012 \\
	\end{tabular}
	\caption{Dataset summary. $L$: the number of labels. $l$: maximum sentence length. $N$: dataset size. $|V|$: vocabulary size. $N_V$: validation set size.}
	\label{tab:data}
\end{table*}

Below we present an example of a doctor's note from our dataset.
Note the use of special characters, abbreviations (hx, s/p, po, pt, \emph{etc.}) and typos.
\begin{quote}
	\textit{\bf Note}: 76 yo woman with hx of cad s/p stent, degen disc disease, ? dementia, ? past tia, htn, and anxiety d/o seen multiple times in the ed for lapses in personal hygeine and ``spells'' now brought in by her daughter after ``hollering and screaming'' this morning at home (where she lives with another daughter). pt denies anger or being upset this morning, has no current complaints. no difficulty with po intake or taking medications, per pt. continues to endorse pain in l lower abdomen yesterday, which is now resolved. pt states that she has had diarrhea x 1 week or so, but no difficulty or pain with urination. no fevers, chills, or other symptoms. no prior hx of similar problem.
	
	\textit{\bf Diagnoses}: anxiety, hypertension.
\end{quote}

It is well understood that diagnoses exhibit a rich correlation (co-occurrence) structure, {\emph e.g.}, anxiety and hypertension as in the example above.
Leveraging such correlations may be very beneficial in practice, when attempting to make multiple predictions for a single patient based solely on doctor's notes.

The dataset has over 50,128 doctor notes and 8,279 free-text diagnoses.
However, $7279$ out of $8279$ diagnoses have less than $10$ instances.
Since a considerable amount of diagnoses do not have enough number of samples to build and reliably evaluate a supervised model, we generate three sub-datasets by focusing on the most common diagnoses.

Table~\ref{tab:data} shows a summary of the datasets being considered.
Different datasets are used for different experiments. 
\textit{ehr25} consists of patients with a single diagnosis, while restricting to the most common $25$ diagnoses (each diagnosis has at least $140$ instances).
\textit{ehr1000} consists of patients with at least one diagnosis, while restricting to the most common 1000 diagnoses (each diagnosis has at least $10$ instances).
\textit{ehr64-all} consists of all doctors' notes including patients with the most common 64 diagnoses ($26071$ instances) and ``control patients'' with no labels ($24057$ instances).
Note that the ``control patients'' group has diagnoses information available but they are not within the top 64 set.

\section{Experiments}\label{sec-exp}
Our experiments are conducted on different subsets of the cohort described in Section~\ref{sc:data}. We compare our convolutional residual models with $n=\{1,2,4,8\}$ residual layers against five baselines on the \textit{ehr1000} dataset.
Then we use the trained parameters of the CNN encoder on the single-labeled \textit{ehr25} dataset to visualize CNN encoded doctor's notes and their corresponding diagnosis using $t$-SNE \cite{t-SNE}.
Finally, we compare the results between a plain CNN classifier and convolutional residual model with $8$ layers on the \textit{ehr64-all} dataset.
Our model aims to encode plain text to sentence vectors using CNNs. However, to the best of the author's knowledge, all of publicly available multi-label text classification datasets have been preprocessed as numerical vectors such as bag-of-words. Besides, generating artificial data for this task is extremely difficult due to the challenging nature of generating synthetic natural language. Therefore, we focus on the real EHR dataset to find the best approach for the intended application task.

For the word embedding, we specify an embedding matrix $W_{\rm emb}\in \mathbb{R}^{k\times|V|}$ for each word in sentence $S$, where $V$ is a fixed-sized vocabulary of size $|V|$, and each word is represented by a $k$-dimensional real-valued word vector.
In \textit{word2vec}, the dimensionality $k$ is set to 300.
For words that do not occur in the \textit{word2vec} embedding matrix, we {\em learn} a corresponding $300$-dimensional word vector (initialized with entries randomly sampled from $[-0.25, 0.25]$, such that the new vectors have approximately the same variance as the pre-trained vectors).
Hence, embeddings are learned for common misspelled words and typos, typically found in doctor notes.
The subset of words in the validation set that do not appear in the training set are replaced by the ``unknown'' token from \textit{word2vec}, because they only appear once in the entire dataset.
This subset accounts for 0.5\% per-sentence words or up to one word per validation sentence.
For the CNN encoder, we set filter windows of sizes $h=\{3,4,5\}$ with $100$ filters each, thus each sentence is encoded as a $300$-dimensional vector.

All models are implemented in Theano \cite{Theano}, using a NVIDIA Titan X GPU with 12 GB memory.
The word embedding matrix is initialized by \textit{word2vec}.
All parameters of the model, namely weights of the CNN encoder and weights in the residual multi-layer classifier are initialized from a uniform distribution with support $[-0.01, 0.01]$.
All bias terms are initialized to zero.
We use Adam \cite{Adam} with learning rate $0.0002$ for the optimization procedure.
$10\%$ of the dataset is used as validation set.
Early stopping is employed to avoid overfitting.
The minibatch size is set to $50$.
We use dropout \cite{srivastava2014dropout} with a probability of $0.50$ for the CNN encoder.
We will make the source codes publicly available.
\begin{figure*}[t!]
	\centering 
	\includegraphics[width=6in]{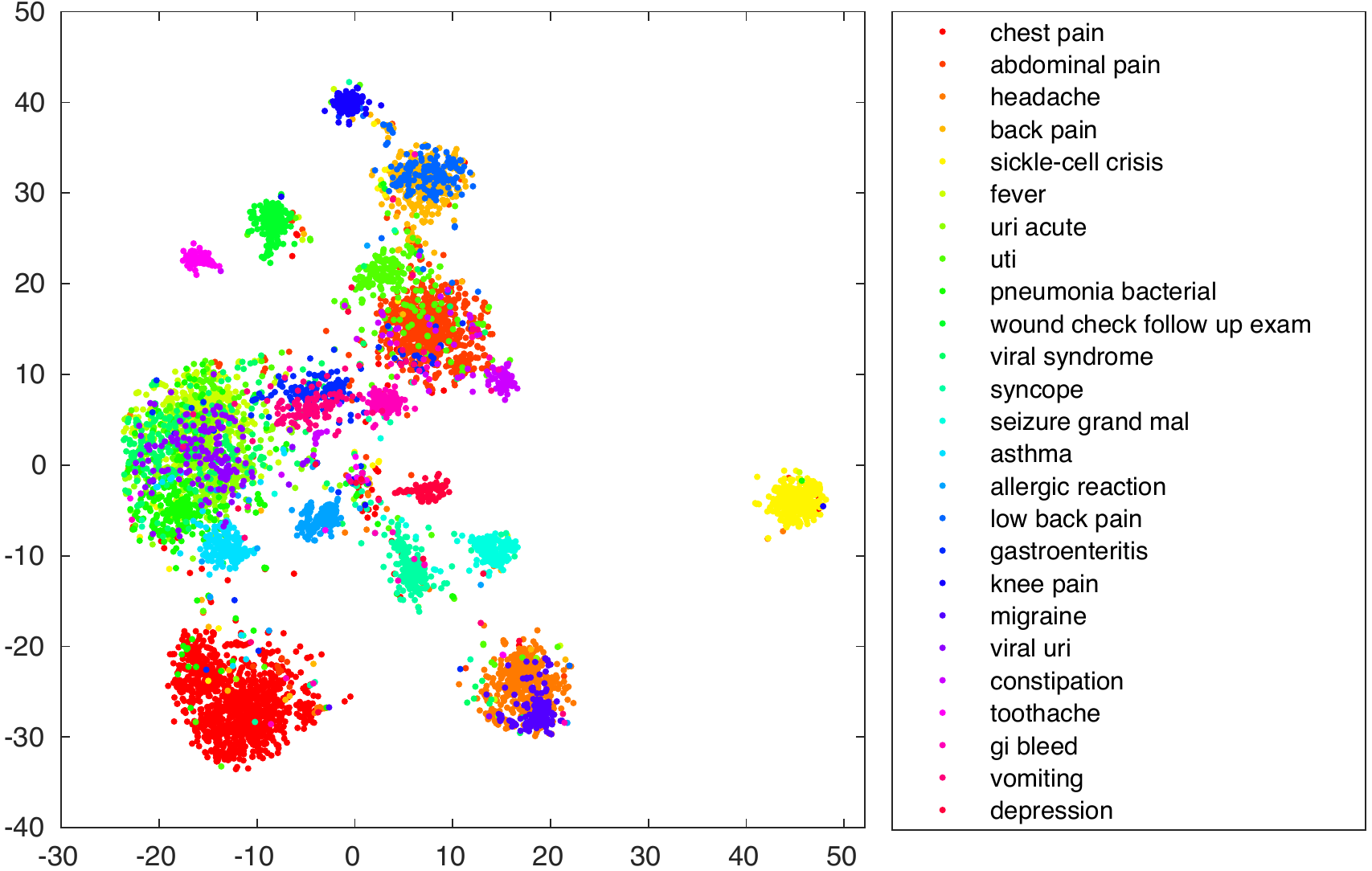} 
	\caption{Visualization of CNN-encoded doctor's notes on \textit{ehr25} using $t$-SNE.}
	\label{fig:visualization}
	%\vspace{-2mm}
\end{figure*}
\subsection{Baselines}

\paragraph{SLEEC} 
is an embedding-based multi-label learning technique proposed by \cite{bhatia2015sparse}. 
This bag-of-words algorithm can be used for large-scale problems by learning embeddings which preserve pairwise distances between nearest label vectors.

\paragraph{FastXML} 
is a tree-based multi-label learning technique proposed by \cite{prabhu2014fastxml}.
This bag-of-words algorithm can make fast predictions by directly optimizing an nDCG-based ranking loss function.

\paragraph{Bi-LSTM}
A Bidirectional LSTM \cite{graves2013hybrid} consisting of paired LSTMs connected in opposite directions is used for sentence encoding.
Then we build the logistic regression model in \eqref{eqt:class} on top of it as classifier.

\paragraph{CNN} 
A logistic regression model in \eqref{eqt:class} on top of the CNN-encoded sentence features as described in Section~\ref{cnnclassifier}.

\paragraph{CRBM}
A conditional restricted Boltzmann machine model as in Section \ref{sec-CRBM}, using CNN-encoded sentence vectors as input features.
The parameters of the CNN encoder are trained by a CNN classifier.

\paragraph{Convolutional Plain Models}
An $8$-layer convolutional plain model (no shortcut connections).
The model has exactly the same set of parameters as an $8$-layer convolutional residual model.
\subsection{Evaluation Metrics}
\paragraph{Precision@$k$}
Precision at $k$ is a popular evaluation metric for multi-label classification problems. Given the ground truth label vector $\yv\in\{0,1\}^L$ and the prediction $\hat{\yv}\in[0,1]^L$, $P@k$ is defined as
\begin{align*}
	P@k:=\frac{1}{k}\sum_{l\in {\rm rank}_k(\hat{\yv})}y^{(l)} \,.
\end{align*}
Precision at $k$ performs a sentence-wise evaluation that counts the fraction of correct predictions in the top $k$ scoring labels. 

\paragraph{nDCG@$k$}
normalized Discounted Cumulative Gain (nDCG) at rank $k$ is a family of ranking measures widely used in multi-label learning. DCG is the total gain accumulated at a particular rank $p$, which is defined as
\begin{align*}
	DCG@k:=\sum_{l\in {\rm rank}_k(\hat{\yv})}\frac{y^{(l)}}{\log(l+1)} \,.
\end{align*}
Then normalizing DCG by the value at rank $k$ of the ideal ranking gives
\begin{align*}
	N@k:=\frac{DCG@k}{\sum_{l=1}^{\min(k,\|\yv\|_0)}\frac{1}{\log(l+1)}}
\end{align*}
Here, nDCG at $k$ ($N@k$) performs a sentence-wise evaluation.

\paragraph{AUC}
The area under the receiving operating characteristic curve (AUC) is the probability that a classifier will rank a randomly chosen positive instance higher than a randomly chosen negative one.
Though originally defined for binary problems, the labels can be represented as a $N\times L$ binary matrix.
We estimate the AUCs of each label individually, then take the average.
AUC is then a label-wise evaluation metric averaged over sentences.
\begin{table*}[t!]
	\centering%\small
	\begin{tabular}{l|r|r|r|r|r|r}
		Model & P@1(\%) & P@3(\%) & P@5(\%) & N@3(\%) & N@5(\%) & AUC(\%) \\ 
		\hline
		SLEEC & 36.36 & 19.74 & 14.03 & 43.00 & 46.64 & --- \\
		fast-XML & 57.64 & 29.38 & 20.04 & 65.66 & 69.52 & 91.84 \\	
		Bi-LSTM & 55.49 & 28.35 & 19.67 & 73.53 & 73.52 & 92.48 \\
		CNN & 58.87 & 30.59 & 20.77 & 78.91 & 78.88 & 93.38 \\
		CRBM & 54.81 & 26.69 & 18.31 & 60.93 & 64.62 & 92.59 \\
		CNN-Plain-8layer & 8.99 & 8.19 & 7.19 & 20.68 & 20.67 & 50.04 \\\hline
		CNN-Res-1layer & 59.98 & 31.04 & 21.11 & 79.71 & 79.67 & 94.56 \\
		CNN-Res-2layer & 60.08 & 31.04 & 21.12 & 79.77 & 79.73 & 94.61 \\
		CNN-Res-4layer & 60.28 & 31.20 & \textbf{21.32} & 80.12 & 80.09 & 94.78 \\
		CNN-Res-8layer & \textbf{60.30} & \textbf{31.21} & 21.27 & \textbf{80.21} & \textbf{80.17} & \textbf{94.89}
	\end{tabular}
	\caption{Quantitative results using several metrics for doctor-notes multi-label classification task on \textit{ehr1000}.  Scores correspond to validation data. The best results are in bold. For SLEEC we do not report an AUC value because the method does not return a classification score.}
	\label{tab:ours}
	%\vspace{-2mm}
\end{table*}

\subsection{Results}
%Multi-label learning evaluation metrics precision at $k$ (P@k), nDCG at $k$ (N@k), and AUC are employed in this work. The quantitative results

\begin{wrapfigure}{R}{0.5\textwidth}	
	\centering
	\includegraphics[width=0.5\textwidth]{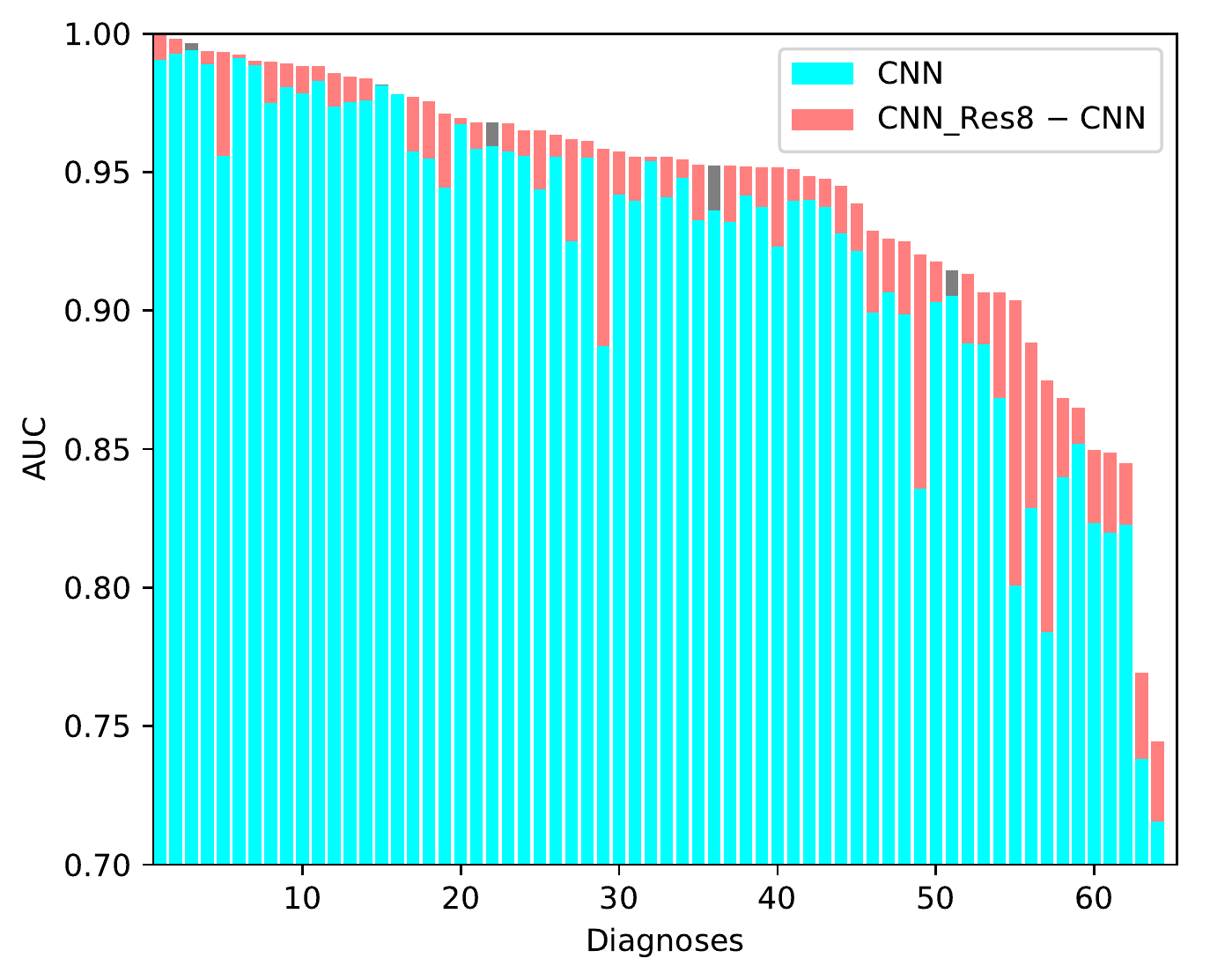}
	\caption{AUC Comparison between a CNN classifier and a 8-layer convolutional residual model on \textit{ehr64-all}. The mean AUC of the CNN classifier and the $8$-layer convolutional residual model are $0.923\pm0.063$ and $0.942\pm0.051$, respectively.}
	\label{fig:CNN_vs_Res8}
\end{wrapfigure}

The precision at $1,3,5$, nDCG at $3,5$, and AUC scores for all methods on the \textit{ehr1000} dataset are shown in Table~\ref{tab:ours}.
Our models show the best multi-label classification performance compared to all other competing methods. The model benefits from adding residual layers in particular when it is shallow. As can be seen, convolutional residual models with $4$ or $8$ residual layers have better quantitative results than those with $1$ or $2$ layers, while the performance difference between models with $4$ layers and $8$ layers is not significant. Moreover, the model converges faster as the number of residual layers increases, e.g., it takes $38$ and $25$ epochs for CNN-Res-$1$layer and CNN-Res-$8$layer to converge, respectively.
Note that the $8$-layer convolutional plain model, which has the same number of parameters as the $8$-layer convolutional residual model, fails completely, indicating that the performance improvement of the proposed model is not simply due to the number of parameters increasing with the number of layers.
Models using CNN-encoded sentence vectors as input features generally perform better than those using bag-of-words representations and LSTM-encoded sentence vectors as input features.
This demonstrates that the CNN encoder is better at capturing sentence structures for the long and noisy doctors' notes in our data.

We use the trained CNN parameters to encode single-label sentences in $ehr25$. In Figure \ref{fig:visualization}, the CNN-encoded sentence vectors (300-dimensional) are projected into a 2-dimensional space using t-SNE.
As can be seen from the visualization plot, the encoded doctors' notes with the same diagnoses are generally clustered together.
Even for those regions containing a mix of colors, they usually have similar diagnoses, such headache and migraine, back pain and low back pain, {\em etc}. 

Finally, we compare the convolutional residual model with $8$ layers against a CNN classifier on the \textit{ehr64-all} dataset, in which about half of doctors' notes are from ``control patients''.
As shown in Figure \ref{fig:CNN_vs_Res8}, our proposed model is not only more accurate on multi-label prediction, but some diagnoses exhibit significant performance gains.
The AUC scores of $59$ (of $64$) diagnoses (red bars) are improved in the $8$-layer residual network compared to the CNN classifier. 
For the remaining 5 (of $64$) diagnoses (dark bars) CNN is only marginally better than the convolutional residual model.
The top 5 diagnoses that increase the most in Figure \ref{fig:CNN_vs_Res8} are hyperglycemia, hypotesion, hyperkalemia, uti, and fever, which are also most likely to have comorbidities in our dataset.
These results demonstrate that our residual model can improve prediction performance by leveraging diagnosis dependencies.
%The top and bottom diagnoses in terms of AUC are $\{$sickle-cell crisis, toothache, ankle sprain, cervical strain, migraine$\}$ and $\{$hypokalemia, tachycardia, renal failure acute, uti, dehydration$\}$, respectively.
%
%\begin{figure}[h!]
%	\centering 
%	\includegraphics[width=2.5in]{foo.pdf} 
%	\caption{AUC Comparison between a CNN classifier and a 8-layer convolutional residual model on \textit{ehr64-all}. The mean AUC of the CNN classifier and the $8$-layer convolutional residual model are $0.923\pm0.063$ and $0.942\pm0.051$, respectively.}
%	\label{fig:CNN_vs_Res8}
%	\vspace{-2mm}
%\end{figure}

\section{Conclusion}
We developed a novel convolutional residual model for multi-label text learning.
A CNN performs convolution and pooling operations to encode the input sentence into a fixed-length feature vector.
Motivated by the local conditional probabilities in CRBM models, we proposed a feedforward neural network with shortcut connections on top of the CNN encoder as classifier.
This classification structure forms a deep residual network to combine information from the encoded sentence vector and label dependencies captured from the label layer.
We presented experiments on a new dataset to predict diagnoses based on plain text from doctor notes.
Experimental results demonstrated the superiority of the proposed model for text embedding and multi-label learning.

% ACKNOWLEDGEMENTS ONLY GO IN THE CAMERA-READY, NOT THE SUBMISSION
% \acks{Many thanks to all collaborators and funders!}

\bibliography{multilabel}

\end{document}